# Image Fusion Techniques in Remote Sensing

Reham Gharbia[1], Ahmad Taher Azar[2], Ali El Baz[3], Aboul Ella Hassanien[4]

[1, 2, 4]Scientific Research Group in Egypt (SRGE)
[2]Faculty of computers and information, Benha University, Egypt.

[3]Faculty of science, Damietta University

[4]Faculty of Computers and Information - Cairo University

{[1]reham_ghrabia@yahoo.com, [2]ahmad_t_azar @ieee.org, [4]aboitcairo@gmail.com}

**Abstract**- Remote sensing image fusion is an effective way to use a large volume of data from multisensor images. Most earth satellites such as SPOT, Landsat 7, IKONOS and QuickBird provide both panchromatic (Pan) images at a higher spatial resolution and multispectral (MS) images at a lower spatial resolution and many remote sensing applications require both high spatial and high spectral resolutions, especially for GIS based applications. An effective image fusion technique can produce such remotely sensed images. Image fusion is the combination of two or more different images to form a new image by using a certain algorithm to obtain more and better information about an object or a study area than. The image fusion is performed at three different processing levels which are pixel level, feature level and decision level according to the stage at which the fusion takes place. There are many image fusion methods that can be used to produce high resolution multispectral images from a high resolution pan image and low resolution multispectral images. This paper explores the major remote sensing data fusion techniques at pixel level and reviews the concept, principals, limitations and advantages for each technique. This paper focused on traditional techniques like intensity hue-saturation-(HIS), Brovey, principal component analysis (PCA) and Wavelet.

*Keywords*-- image fusion, pan-sharpening, Brovey transform, PCA, IHS, and Wavelet transform.

I. INTRODUCTION

In remote sensing and mapping, there are many applications that simultaneously require the high spatial and high spectral resolution in a signal image such as classification, change detection and land cover, increased need for image fusion. The fused image usually has more information about the target or scene than any of the individual images used in the fusion process. The objective of information fusion is to improve the accuracy of image interpretation and analysis by making use of complementary information [1-3]. Data fusion has been widely used in remotely sensed image analysis at pixel, feature, and decision level [4-7]. Images used for fusion can be taken from multimodal imaging sensors or from the same imaging sensor at different times. There are several benefits in using image fusion: wider spatial and temporal coverage, decreased uncertainty, improved reliability, and increased robustness of system performance [8]. The basic problem of image fusion is one of determining the best procedure for combining the multiple input images. Many image fusion techniques have been developed to merge a Pan image and a MS image into a multispectral image with high spatial and spectral resolution simultaneously. An ideal image fusion technique should have three essential factors, i.e. high computational efficiency, preserving high spatial resolution and reducing color distortion [9]. In the past few years, many image fusion methods have been proposed, such as intensity hue-saturation- (IHS), Brovey transform (BT), principal component analysis (PCA) and wavelets.

II. IMAGE FUSION TECHNIQUES

In this section, we will briefly introduce various types of image fusion techniques. The main principle, and typical advantages and disadvantages are introduced for each approach. Image fusion techniques include for example not for all

A. *IHS image fusion technique*

The classical image fusion techniques include intensity-hue-saturation transform technique (IHS) [10]. IHS is a common way of fusing high spatial resolution, single band, pan image and low spatial resolution, multispectral remote sensing image. The R, G and B bands of the multispectral image are transformed into IHS components, replacing the intensity component by the pan image, and performing the inverse transformation to obtain a high spatial resolution multispectral image [10, 11] (see Fig. 1). IHS can enhance spatial details of the multispectral image and improve the textural characteristics of the fused image, but the fusion image exist serious spectral distortion [12]. The IHS transform is used for geologic mapping because the

IHS transform could allow diverse forms of spectral and spatial landscape information to be combined into a single data set for analysis [13].

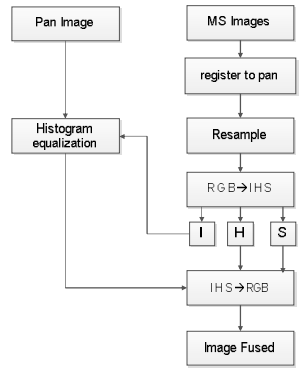

Fig.1 scheme of IHS image fusion

Although the IHS method has been widely used, the method cannot decompose an image into different frequencies in frequency space such as higher or lower frequency. Hence the IHS method cannot be used to enhance certain image characteristics [14]. The color distortion of IHS technique is often significant. To reduce the color distortion, the PAN image is matched to the intensity component before the replacement or the hue and saturation components are stretching before the reverse transform. Also propose a method that combines a standard IHS transform with FFT filtering of both the pan image and the intensity component of the original multispectral image to reduce color distortion in the fused image [15]. Schetselaar modified IHS and presented a new method that preserves the spectral balance of the multispectral image data and modulates the IHS coordinate uniformly [16]. The method takes the limits in the representation of color of the display device into account, which aids in compromising the amount and spatial distribution of the over-range pixels against contrast in intensity and saturation. There are other improvements about IHS such as using wavelet [17-19]. Also, Image fusion based on the nonsubsampled Contourlet transform (NSCT) and IHS achieved increased in retaining the spectral information and spatial details, and better integration effect [20]. With IHS transform, the segment based fusion was developed specifically or a spectral characteristics preserving image merge coupled with a spatial domain filtering [21]. With experiment on SPOT and TM remote sensing images demonstrates that the IHS fusion method can improve the fused result with a viewpoint of synthetically quality evaluation. It is a practical and effective method for these two types of images [22].

*B. The Brovey transform image fusion technique*

The BT uses a mathematical combination of the MS bands and PAN band. Each MS band is multiplied by a ratio of the PAN band divided by the sum of the MS bands. The fused R, G, B bands [23]. Many researchers used the BT to fuse a RGB image with a high resolution image [24-27]. The BT image fusion is used to combine Landsat TM and radar SAR (ERS-1) images [28] and successful for Spot pan fusion [29]. The BT is limited to three bands and the multiplicative techniques introduce significant radiometric distortion. In addition, successful application of this technique requires an experienced analyst [29] for the specific adaptation of parameters. This precludes development of a user-friendly automated tool.

*C. The Principle Component Analysis (PCA) image fusion technique*

PCA transformation is a technique from statistics for simplifying a data set. It was developed by Pearson 1901 and Hotelling 1933, whilst the best modern reference is Jolliffe, 2002. The aim of the method is to reduce the dimensionality of multivariate data whilst preserving as much of the relevant information as possible. It translates correlated data set to uncorrelated dataset. By using this method, the redundancy of the image data can be decreased [3]. The multi-spectral image is transformed with PCA transform, and the eigenvalues and corresponding eigenvectors of correlation matrix between images in the multi-spectral image's individual bands are calculated out to obtain each matrix's principle components. The first principle component of the multi-spectral image is replaced with the matched pan image, and then we get the new first principle component. The new first principle component and other principle components are transformed with inverse PCA transform to form the fused image. We replace the first principal component image with pan image data because the first principle component image has the common information to all the bands [30] (see Fig. 2). The traditional PCA fusion method may not be satisfactory to fuse high-resolution images and low-resolution multi-spectral images, because it may distort the spectral characteristics of the multi-spectral data [31]. The adaptive PCA approach provides efficient spectral transformation between the two images by selecting the best principal component to be replaced by the Pan image with high spatial resolution [32] Singha developed a PCA based image fusion technique in face recognition using both thermal and visible images [33]. The main advantage of PCA is that you are able to have a large number of inputs and that most of the information within all the inputs can be compressed into a much smaller amount of outputs without much loss of information [34].

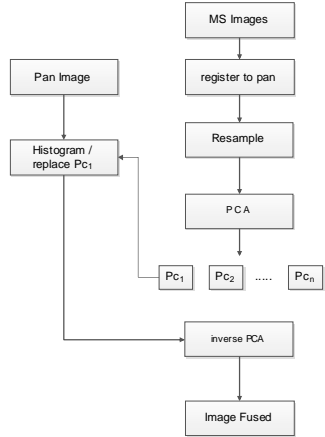

Fig.2 scheme of PCA image fusion

One disadvantage of the use of PCA for image fusion is that you are selecting only the first eigenvector to describe your data set. Even though this eigenvector contains 90% of the shared information there is still some information that will not be evident in the final fused image. PCA performs well when compared to other image fusion techniques. PCA was superior to the simple averaging technique and the Morphological pyramid technique for the majority of the measures and only being inferior to the discrete wavelet transform [35].

D. *The wavelet image fusion technique*

Wavelet theory is related to multi-resolution analysis theory. Since it was invented in 1980, many theoretical and application researches have been conducted. In recent years, wavelet transform has been introduced in image fusion domain [36-43], due to its multi-resolution analysis (MRA) characteristic [44, 45]. The traditional wavelet-based image fusion can be performed by decomposed the two input images separately into approximate coefficients and detailed coefficients then high detailed coefficients of the multi-spectral image are replaced with those of the pan image. The new wavelet coefficients of the multi-spectral image are transformed with the inverse wavelet transform to obtain the fusion multi-spectral image (see Fig. 3). The wavelet image fusion technique can improve the spatial resolution while preserve the spectral characteristics at a maximum degree [46].but the method discards low frequency component of the pan image completely, it may produce mosaic for the fusion image. When the images are smooth, without abrupt intensity changes, the wavelets work appropriately, improving the results of the classical methods. This has been verified with smooth images and also with medical images, where no significant changes are present.

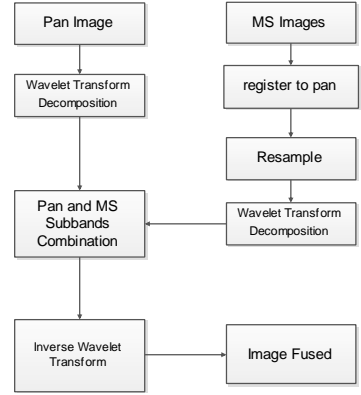

Fig.3 scheme of Wavelet image fusion

In this case, the type of images (remote sensing, medical) is irrelevant [47]. Recently more studies propose hybrid schemes, which use wavelets to extract the detail information from one image and standard image transformations to inject it into another image, or propose improvements in the method of injecting information [48-.55] These approaches seem to achieve better results than either the standard image fusion schemes e.g. (IHS, than either the standard image fusion schemes (e.g. IHS, PCA) or standard wavelet-based image fusion schemes (e.g. substitution, addition); however, they involve greater computational complexity.

III. ANALYSIS AND DISCUSSION

The BT is limited to three bands and the multiplicative techniques introduce significant radiometric distortion. In addition, successful application of this technique requires an experienced analyst. The traditional PCA fusion method may not be satisfactory to fuse high-resolution images and low-resolution multi-spectral images, because it may distort the spectral characteristics of the multi-spectral data. The main advantage of PCA is that you are able to have a large number of inputs and that most of the information within all the inputs can be compressed into a much smaller amount of outputs without much loss of information. The wavelet image fusion is wide used in remote sensing due to its multi-resolution analysis (MRA) characteristic. The wavelet image fusion technique can improve the spatial resolution while preserve the spectral characteristics at a maximum degree.
We can summarize the advantage and disadvantage for each technique in Table 1

| Tech. | Advantages | Disadvantages |
|---|---|---|
| BT | -low computational demand<br>-decreased time<br>-simple<br>-used as benchmark for comparison of other fusion methods | -high contrast pixel values in input<br>-image are depressed in value in the fused image<br>-this method does not give guarantee to have a clear objects from the set of images. |
| PCA | -selects optimal weighting coefficients based on information content<br>-removes redundancy present in input image<br>-compress large amounts of inputs without much loss of information | -usually selects 1st Eigen value which does not contain all of the patterns between inputs<br>-fused image will be of lesser quality than any of the input images<br>-strong correlation between the input images and fused image is needed |
| WT | -different rules are applied to the low and high frequency portions of the signal<br>-Performs favourably when compared to other fusion methods | -is not shift-invariant<br>-pixel by pixel analysis is not possible<br>-not possible to fuse images of different sizes<br>-In this technique final fused image have a less spatial resolution. |
| IHS | -the intensity change has little effect on the spectral information and is easy to deal with | -IHS destroys the spectral characteristics of the MS data.<br><br>The large difference between the values of the PAN and intensity images appears to cause the large spectral distortion of fused MS images. Indeed, this difference (PAN-I) causes the altered saturation component in the RGB-IHS conversion model. |

## IV. CONCLUSION

Hybrid schemes, and advanced schemes. In all cases, the authors found that their proposed methods performed well relative to a selection of other methods. While some schemes have clear advantages over other schemes, such as faster processing time, lower complexity, or lower memory requirements, it is not possible to draw absolute conclusions about which are the best or worst without first performing extensive tests on all of the schemes. Most of the above schemes were tested on SPOT imagery, while only a few were tested on IKONOS or QuickBird imagery or on a fusion of SPOT and Landsat imagery. Since the sensors have different characteristics, particularly in terms of the sensitivity range of the pan sensor, the experimental results do not necessarily represent typical results. Furthermore, the types of regions contained in the test imagery also affect results. Some schemes might perform better on vegetation while others might perform better on urban areas, and this might not be reflected in the experimental results presented. Finally, the application for which the fused image is needed determines which quality factors are most important. In some cases, the highest visual quality might be essential, while in other cases high spectral or spatial correlation might be more desirable.

Remote sensing image fusion is an effective way to use the large volume of data from multisource images. And it can combine multisensor, multi-temporal, multispectral and multiresolution images for analyzing. Thus it can overcome the problem of information deficiency during artificial extraction of remote sensing images. In recent years, there are many research works on methods of image fusion such as, IHS image fusion technique is wide used in remote sensing fusion because IHS can enhance spatial details of the multi-spectral image and improve the textural characteristics of the fused image, but the fusion image exist serious spectral distortion. We can overcome this problem by improving IHS through integration with other techniques such as wavelet. With experiment on SPOT and TM remote sensing images demonstrates that the IHS fusion method can improve the fused result with a viewpoint of synthetically quality evaluation. It is a practical and effective method for these two types of images.

We conclude that the traditional image fusion techniques have limitation and do not meet the needs of remote sensing therefore our way is the only hybrid systems. Hybrid techniques in pixel level are more efficiency techniques than traditional techniques. Therefore, we are heading in our future work on improving conventional systems by working on hybrid systems.